\documentclass[10pt,twocolumn,letterpaper]{article}

\usepackage[pagenumbers]{cvpr} % To force page numbers, e.g. for an arXiv version

\usepackage{graphicx}
\usepackage{amsmath}
\usepackage{amssymb}
\usepackage{booktabs}

\usepackage{multirow}
\usepackage{color}

\usepackage[pagebackref,breaklinks,colorlinks]{hyperref}

\usepackage{mathtools}

\usepackage{bbm}
\usepackage[capitalize]{cleveref}
\crefname{section}{Sec.}{Secs.}
\Crefname{section}{Section}{Sections}
\Crefname{table}{Table}{Tables}
\crefname{table}{Tab.}{Tabs.}

\def\onedot{\ifx\@let@token.\else.\null\fi\xspace}
\def\eg{\emph{e.g}\onedot} 
\def\ie{\emph{i.e}\onedot}

\newcommand{\Tref}[1]{Table~\ref{#1}}
\newcommand{\Eref}[1]{Eq.~(\ref{#1})}
\newcommand{\Fref}[1]{Fig.~\ref{#1}}

\newcommand{\Sref}[1]{Sec.~\ref{#1}}

\begin{document}

\title{Generative Bias for Robust Visual Question Answering}

\author{Jae Won Cho$^{1}$ \quad
Dong-Jin Kim$^{2}$ \quad
Hyeonggon Ryu$^{1}$ \quad
In So Kweon$^{1}$\\
\\
$^{1}$KAIST \quad $^{2}$Hanyang University\\
% \small{
%   \textsuperscript{1}\texttt{\{???, ????, iskweon\}@kaist.ac.kr}\qquad
%   \textsuperscript{2}\texttt{\href{mailto:djdkim@hanyang.ac.kr}{\color{black} djdkim@hanyang.ac.kr}}
%   }\\
}
\maketitle

%%%%%%%%% ABSTRACT
\begin{abstract}
   The task of Visual Question Answering (VQA) is known to be plagued by the issue of VQA models exploiting biases within the dataset to make its final prediction. Various previous ensemble based debiasing methods have been proposed where an additional model is purposefully trained to be biased in order to train a robust target model. However, these methods compute the bias for a model simply from the label statistics of the training data or from single modal branches. In this work, in order to better learn the bias a target VQA model suffers from, we propose a generative method to train the bias model \emph{directly from the target model}, called GenB. In particular, GenB employs a generative network to learn the bias in the target model through a combination of the adversarial objective and knowledge distillation. We then debias our target model with GenB as a bias model, and show through extensive experiments the effects of our method on various VQA bias datasets including VQA-CP2, VQA-CP1, GQA-OOD, and VQA-CE, and show state-of-the-art results with the LXMERT architecture on VQA-CP2.
\end{abstract}
\vspace{-5mm}

\section{Introduction}
\label{sec.intro}

Visual Question Answering (VQA)~\cite{antol2015vqa} is a challenging multi-modal task that requires a model to correctly understand and predict an answer given an input pair of image and question.
Various studies have shown that VQA is prone to biases within the dataset and tend to rely heavily on language biases that exists within the dataset~\cite{agrawal2016analyzing,goyal2017making,zhang2016yin}, and VQA models tend to predict similar answers only depending on the question regardless of the image. 
In response to this, recent works have developed various bias reduction techniques, and recent methods have exploited ensemble based debiasing methods~\cite{cadene2019rubi,clark2019dont,han2021greedy,ramakrishnan2018overcoming} extensively.

\begin{figure}
    \centering
    \includegraphics[width=1\linewidth]{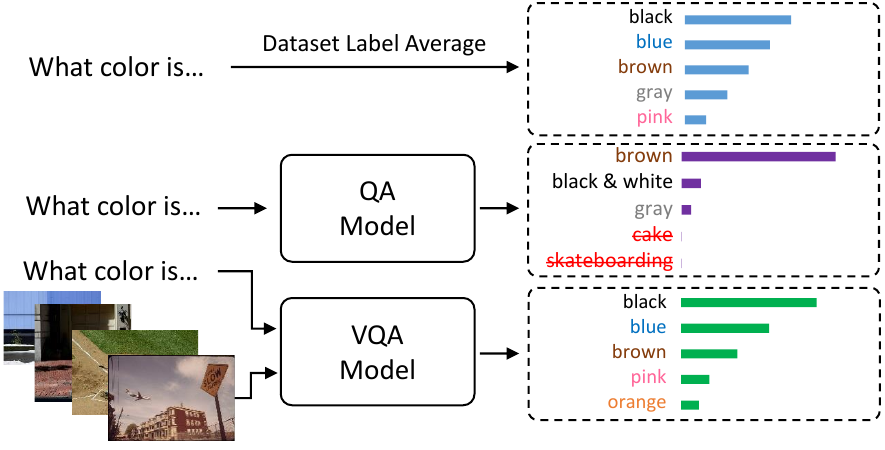}
    \caption{Given a Question Type (``What color is...''), we show all of the averaged answers within the training dataset. 
    The answer computed from the entire training dataset is the known dataset label average or dataset bias as in~\cite{clark2019dont,han2021greedy}. We see that the averaged model predictions of the Question-Answer Model and Visual-Question-Answer Model are significantly different. 
    }
    \vspace{-5mm}
    \label{fig:teaser}
\end{figure}

Among ensemble based methods, additional models are introduced to concurrently learn biases that might exist within each modality or dataset. For example, in works such as~\cite{cadene2019rubi,han2021greedy}, the Question-Answer model is utilized to determine the language prior biases that exist when a model is asked to give an answer based solely off of the question. 
This Question-Answer model is then utilized to train a robust ``target'' model, which is used for inference. 
The key purpose of an ensemble ``bias'' model is to capture the biases that are formed with its given inputs (\ie, language prior biases from the Question-Answer model).
In doing so, if this model is able to represent the bias well, this bias model can be used to teach the target model to avoid such biased answers. 
In other words, the better the bias model can learn the biases, the better the target model can avoid such biases.

Existing ensemble based methods either use pre-computed label statistics of training data (GGE-D~\cite{han2021greedy} and LMH~\cite{clark2019dont}), or single modal branches 
that compute
the answer 
from either the question or image~~\cite{cadene2019rubi,clark2019dont,han2021greedy,niu2021cfvqa}. 
However, we 
conjecture that there is a limit to the bias representation that can be obtained from such methods,
as the model's representative capacity is limited by its inputs. 
In addition, pre-computed label statistics represents only part of the bias~\cite{han2021greedy}.
As shown in~\Fref{fig:teaser}, 
given a question type, the pre-computed label statistics (or known dataset bias) are noticeably different to the predictions of a model trained with the question or with the image and question. This discrepancy signifies that there is a part of the bias that we cannot fully model simply with the previous methods. 
Therefore, we propose a novel stochastic bias model that learns the bias \emph{directly from the target model}.

More specifically, to directly learn the bias distribution of the \emph{target model}, we model the bias model as a Generative Adversarial Network (GAN)~\cite{goodfellow2014generative} to stochastically mimic the target model's answer distribution given the same question input by introducing a random noise vector.
As seen through literature, most biases are held within the question~\cite{agrawal2016analyzing}, so we use questions as the main bias modality.
To further enforce this, we utilize knowledge distillation~\cite{hinton2015distilling} on top of adversarial training to force the bias model to be as close as possible to the target model, 
so that the target model learns from harder negative supervision from the bias model.
Finally, with our generative bias model, we then use our modified debiasing loss function to train our target model.
Our final bias model is able to train the target model that outperforms previous uni-modal and multi-modal ensemble based debiasing methods by a large margin.
To the best of our knowledge, we are the first to train the bias model by directly leveraging the behavior of the target model using a generative model for VQA. 

To show the efficacy and robustness of our method, we perform extensive experiments on commonly used robustness testing VQA datasets and various different VQA architectures.
Our method show the state-of-the-art results on all settings without the use of external human annotations or dataset reshuffling methods.

Our contributions are as follows:
\begin{itemize}
    \item We propose a novel bias model for ensemble based debiasing for VQA by directly leveraging the target model that we name \emph{GenB}.
    \item In order to effectively train GenB, we employ a Generative Adversarial Network and knowledge distillation loss to capture both the dataset distribution bias and the bias from the target model.
    \item We achieve state-of-the-art performance on VQA-CP2, VQA-CP1 as well as the more challenging GQA-OOD dataset and VQA-CE using the simple UpDn baseline without extra annotations or dataset reshuffling and state-of-the-art VQA-CP2 peformance on the LXMERT backbone.
\end{itemize}

\section{Related Work}
\label{sec.related}

VQA~\cite{antol2015vqa,gurari2018vizwiz,johnson2017clevr} has been actively studied in recent years with performance reaching close to human performance~\cite{kim2018bilinear,lxmert2019,vlbert2019,lu2019vilbert,chen2020uniter} in the most recent works.
Even still, the VQA dataset has been notorious for its reliance on language biases as shown by~\cite{agrawal2018don}.
To this date, many VQA datasets and evaluation protocols have been derived from the original VQA dataset and have been released to the public as a means to test and understand the biases and robustness of VQA models such as VQA-CP2 and CP1~\cite{agrawal2018don}, GQA-OOD~\cite{kervadec2021gqaood}, and VQA-CE~\cite{dancette2021beyond}.

In light of this, many methods have been proposed. One line of work strives to improve visual attention through visual grounding by including additional human annotations or explanations~\cite{selvaraju2019taking,wu2019self}, which show limited improvements. Another line of work changes the distribution of the training set by either randomly or systematically replacing the images and questions~\cite{teney2020value,zhu2021overcoming,wen2021debiased,chen2022kddaug,wu2022overcoming} or by augmenting the dataset through counterfactual samples~\cite{chen2020counterfactual} or through the use of an external inpainting network~\cite{gokhale2020mutant}. Although powerful, according to~\cite{niu2021cfvqa}, changing the distribution of the training set does not agree with the philosophy of the creation of this dataset. Therefore, we do not directly compare experimentally with this line of works. Most recently, another line of work has been released where an additional objective of Visual Entailment is added to further boost performance~\cite{si2021check}, which we do not compare for fairness. Another recent work focuses on distilling knowledge from the some of the aforementioned methods to train a robust student model~\cite{niu2021introd}. As this method is built upon other methods, we compare to this method where possible.

One of the most effective line of work is ensemble based methods~\cite{Abbasnejad_2020_CVPR,cadene2019rubi,niu2021cfvqa,han2021greedy} and we place our line of work here. Within ensemble based methods, AReg~\cite{Abbasnejad_2020_CVPR}, RUBi~\cite{cadene2019rubi}, LMH~\cite{clark2019dont}, GGE~\cite{han2021greedy} tackle the language prior directly and only use a question-only model. Unlike these methods, GGE~\cite{han2021greedy} shows the possibility of using the same model as the target model to learn biases but to a limited extent. On the other hand, CF-VQA~\cite{niu2021cfvqa} uses both question and image, but uses the modalities individually without combining them. 
Our work is distinct from all previous ensemble based methods as we use a generative network with a noise input to aid the bias model in learning the bias directly from the target model.
We also believe that our work can be applied to other multi-modal~\cite{jang2022signing,jang2023self,kim2019dense,kim2020detecting,kim2021acp++,kim2021dense,argaw2022anatomy} and uni-modal research~\cite{oh2022daso} tasks in the future.

\section{Methodology}
\label{sec.method}

In this section, we explain VQA briefly and describe in detail our method GenB.

\begin{figure*}[t]
    \centering
    \includegraphics[width=.95\linewidth]{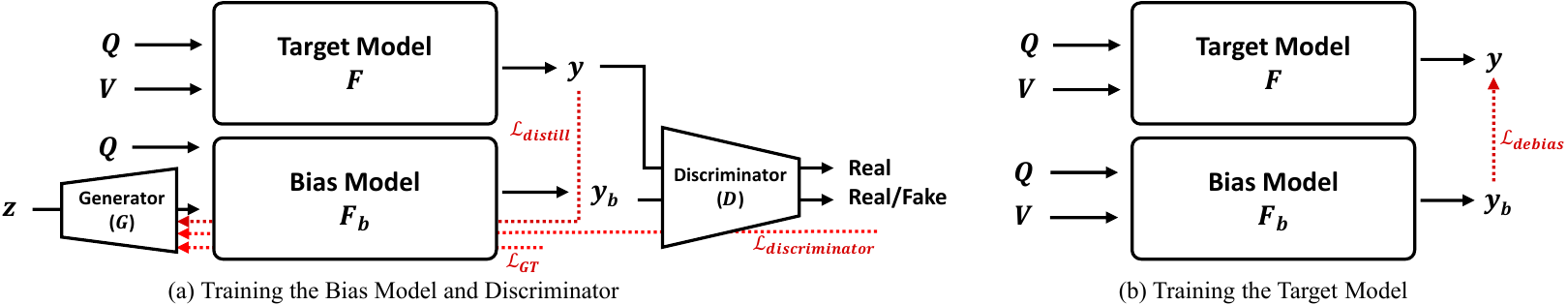}
    \caption{(a) shows how we train our Bias Model and Discriminator. The Bias Model is trained with 3 different losses including the ground truth BCE (\Eref{eq:bceloss}), knowledge distillation (\Eref{eqn:kl}), and adversarial \Eref{eq:ganloss}) losses.
    (b) shows how our Target Model is trained with the Bias model with debiasing loss functions (refer to existing works). Note, steps (a) and (b) happen concurrently and note that we only use the Target Model during inference.
    }
    \vspace{-4mm}
    \label{fig:architecture}
\end{figure*}

\subsection{Visual Question Answering Baseline}

With an image and question as a pair of inputs, a VQA model learns to correctly predict an answer from the whole answer set $\mathcal{A}$. 
A typical VQA model ${F(\cdot, \cdot)}$ takes both a visual representation ${\textbf{v}}\in \mathbb{R}^{n \times d_v}$ 
(a set of feature vectors computed from a Convolutional Neural Network given an image where $n$ is the number of objects in an image and $d_v$ being the vector dimension) and a question representation ${\textbf{q}}\in \mathbb{R}^{d_q}$ (a single vector computed from a GloVe~\cite{pennington2014glove} word embedding followed by a Recurrent Neural Network given a question) as input.
Then, an attention module followed by a multi-layer perceptron classifier ${F : \mathbb{R}^{n\times d_v} \times \mathbb{R}^{d_q} \xrightarrow{} \mathbb{R}^{|\mathcal{A}|}}$ which
generates an answer logit vector $\mathbf{y}\in \mathbb{R}^{|\mathcal{A}|}$ (\ie,~$\mathbf{y} = {F(V, Q)}$). 
Then, after applying a sigmoid function $\sigma(\cdot)$, our goal is to make an answer probability prediction $\sigma(\mathbf{y})\in[0,1]^{|\mathcal{A}|}$ close to the ground truth answer probability $\mathbf{y}_{gt}\in[0,1]^{|\mathcal{A}|}$.
In this work, we adopt one of the popular state-of-the-art architectures UpDn~\cite{anderson2018bottom} that is widely used in VQA debiasing research.

\subsection{Ensembling with Bias Models}

In this work, our scope is bias mitigation through ensembling bias models 
similar to previous works~\cite{cadene2019rubi,clark2019dont,han2021greedy}.
In ensemble based methods, there exist a ``bias'' model that generates $\mathbf{y}_b\in \mathbb{R}^{|\mathcal{A}|}$, which we define as ${F}_b(\cdot, \cdot)$, and a ``target'' model, defined as ${F(\cdot, \cdot)}$. 
Note that, we discard ${F}_b(\cdot, \cdot)$ during testing and only use ${F(\cdot, \cdot)}$.
As previously mentioned, the goal of the existing bias models is to overfit to the bias as much as possible.
Then, given the overfitted bias model, the target model is trained with a debiasing loss function~\cite{cadene2019rubi,clark2019dont,han2021greedy}
to improve the robustness of the target model.
Ultimately, the target model learns to predict an unbiased answer by avoiding the biased answer from the bias model.
The bias model ${F}_b(\cdot, \cdot)$ can either be the same or different from the original ${F}(\cdot, \cdot)$ and there could be multiple models as well~\cite{niu2021cfvqa}. 
Although previous works try to leverage the bias from the individual modalities~\cite{cadene2019rubi,clark2019dont,han2021greedy,niu2021cfvqa},
we propose that this limits the ability of the model to represent biases. 
Hence, in order to represent the biases \emph{similar to the target model}, 
we set the architecture of ${F}_b(\cdot, \cdot)$ to be the same as ${F}(\cdot, \cdot)$ and we use the UpDn~\cite{anderson2018bottom} model.

\subsection{Generative Bias}
\label{sec:GenB}

As mentioned in the \Sref{sec.intro}, as our goal is to train a bias model that can generate stochastic bias representations,
we use a random noise vector in conjunction with a given modality to learn both the dataset bias and the bias that the target model could exhibit.
As the question is known to be prone to bias, we keep the question modality and use it as the input to our bias model ${F}_b(\cdot, \cdot)$. But instead of using the image features,
we introduce a random noise vector $\mathbf{z}\in\mathbb{R}^{n \times 128}$ in addition to a generator network $G : \mathbb{R}^{n\times 128}\xrightarrow{} \mathbb{R}^{n\times d_v}$ to generate the corresponding input to the bias model ${F}_b(\cdot, \cdot)$. 
Formally, given a random Gaussian noise vector $\mathbf{z} \sim \mathcal{N}(\mathbf{0},\mathbf{1})$, a generator network ${G}(\cdot)$ synthesizes a vector that has the same dimension as the image feature representation, \ie,~$\hat{\mathbf{v}} = {G}(\mathbf{z})\in \mathbb{R}^{n \times d_v}$.
Ultimately, our model takes in the question $\mathbf{q}$ and $G(\mathbf{z})$ as its input and generates the bias logit $\mathbf{y_b}$ in the form $F_{b}({G}(\mathbf{z}),\mathbf{q}) = \mathbf{y_b}$. Note, this can be done on another modality, 
(\ie,
$F_{b}({G}(\mathbf{z}),\mathbf{v}) = \mathbf{y_b}$), but we found this is unhelpful. 
For simplicity, we consider generator and bias model as one network and rewrite $F_b(G(\mathbf{z}),\mathbf{q})$ in the form $F_{b,G}(\mathbf{z},\mathbf{q})$ and call our ``Generative Bias'' method \textbf{GenB}.

\subsection{Training the Bias Model}
\label{sec:training}

In order for our bias model GenB to learn the biases given the question, we use the traditional VQA loss, the Binary Cross Entropy Loss:
\begin{equation}
    \mathcal{L}_{GT}(F_{b,G}) = \mathcal{L}_{BCE}(\sigma(F_{b,G}(\mathbf{z},\mathbf{q})), \mathbf{y}_{gt}).
\label{eq:bceloss}
\end{equation}
However, unlike existing works, 
we want the bias model to also capture \emph{the biases in the target model}.
Hence, in order to mimic the bias of the target model as a random distribution of the answers, we propose adversarial training~\cite{goodfellow2014generative} similar to~\cite{kim2019image} to train our  bias model.
In particular, we introduce a discriminator that distinguishes the answers from the target model and the bias model as ``real'' and ``fake'' answers respectively.
The discriminator is formulated as
$D(F(\mathbf{v},\mathbf{q}))$ and $D(F_{b,G}(\mathbf{z},\mathbf{q}))$ or rewritten as
$D(\mathbf{y})$ and $D(\mathbf{y}_b)$.
The objective of our adversarial network with generator $F_{b,G}(\cdot,\cdot)$ and $D(\cdot)$
can be expressed as follows:

\begin{equation}
\begin{split}
    &\min_{F_{b,G}} \max_{D} \mathcal{L}_{GAN}(F_{b,G},D),  \text{where}\\
    &\mathcal{L}_{GAN}(F_{b,G},D) \\
    =&    \mathop{\mathbb{E}}\limits_{{\mathbf{v}, \mathbf{q}}}\Big[\log\Big(D\big(F(\mathbf{v}, \mathbf{q})\big)\Big)\Big]  \\
    &+ \mathop{\mathbb{E}}\limits_{{\mathbf{q},\mathbf{z}}}\Big[\log\Big(1-D\big(F_{b,G}(\mathbf{z}, \mathbf{q})\big)\Big)\Big]
    \\
    =&   {\mathbb{E}}_{{\mathbf{y}}}\Big[\log\big(D(\mathbf{y})\big)\Big] 
    + 
    {\mathbb{E}}_{{\mathbf{y}_b}}\Big[\log\big(1-D(\mathbf{y}_b)\big)\Big].
\end{split}
\label{eq:ganloss}
\end{equation}
The generator ($F_{b,G}$) tries to minimize the objective ($\mathcal{L}_{GAN}$) against an adversarial discriminator ($D$) that tries to maximize it.
Through alternative training of $D$ and $F_{b,G}$, the distribution of the answer vector from the bias model ($\mathbf{y}_b$) should be close to that from the target model ($\mathbf{y}$).

In addition, to further enforce bias model to capture the intricate biases present in the target model, 
we add an additional knowledge distillation objective~\cite{hinton2015distilling} similar to~\cite{kim2018disjoint,kim2020detecting,cho2021dealing,kim2021acp++,kim2021single} to ensure that the model bias model is able to follow the behavior of the target model with only the $\mathbf{q}$ given to it.
We empirically find that it is beneficial to include a sample-wise distance based metric such as KL divergence.
This method is similar to the approaches in the image to image translation task~\cite{isola2017image}.
Then, the goal of the generator is not only to fool the discriminator but also to try to imitate the answer output of the target model in order to give the target model more challenging supervision in the form of \emph{hard negative} sample synthesis. 
We add another objective to our adversarial training for $F_{b,G}(\cdot,\cdot)$:
\begin{equation}
    \mathcal{L}_{distill}(F_{b,G}) = \mathop{\mathbb{E}}\limits_{{\mathbf{v}, \mathbf{q},\mathbf{z}}}\Big[D_{KL}\big(F(\mathbf{v},\mathbf{q}) \rVert {F_{b,G}(\mathbf{z},\mathbf{q})}\big)\Big].
    \label{eqn:kl}
\end{equation}
Ultimately, the final training loss for the bias model, or GenB, is as follows:
\begin{equation}
    \begin{aligned}
    &\min_{F_{b,G}} \max_{D} \mathcal{L}_{GenB}(F_{b,G},D),  \text{where}\\
    &\mathcal{L}_{GenB}(F_{b,G},D) =\\
    & \mathcal{L}_{GAN}(F_{b,G},D)+ \lambda_1 \mathcal{L}_{distill}(F_{b,G}) + \lambda_2 \mathcal{L}_{GT}(F_{b,G}),
    \end{aligned}
    \label{eqn:genb}
\end{equation}
where $\lambda_1$ and $\lambda_2$ are the loss weight hyper-parameters.

\subsection{Debiasing the Target Model}
\label{sec:debiasing}

Given a generated biased answer $\mathbf{y}_b$, there are several debiasing loss functions that can be used such as~\cite{cadene2019rubi,clark2019dont,han2021greedy}, and we show the effects of each one in~\Tref{table:lossabla}.
The GGE~\cite{han2021greedy} loss is one of the best performing losses without the use of label distribution. The GGE loss takes the bias predictions/distributions and generates a gradient in the opposite direction to train the target model.
With this starting point, we modify this equation with the ensemble of the biased model GenB in this work as follows: 
\begin{equation}
    \mathcal{L}_{target}(F) = \mathcal{L}_{BCE}(\mathbf{y}, \mathbf{y}_{DL} ),
    \label{eqn:gradient}
\end{equation}
where the $i$-th element of the pseudo-label $\mathbf{y}_{DL}$ is defined as follows:
\begin{equation}
    \mathbf{y}_{DL}^i = \min\big(1~,~2 \cdot \mathbf{y}_{gt}^i \cdot \sigma(-2 \cdot \mathbf{y}_{gt}^i \cdot \mathbf{y}_{b}^i) \big),
\end{equation}
where $\mathbf{y}_{gt}^i$ and $\mathbf{y}_b^i$ are the $i$-th element of the ground truth and the output of the biased model respectively. % with  being the output of the target model.
The key point of difference is that unlike \cite{han2021greedy} that suppresses the output of the biased model with the sigmoid function, we use $\mathbf{y}_b$ without using the sigmoid function. In this case, as the value of $\mathbf{y}_{DL}$ can exceed $1$, we additionally clip the value so that the value of $\mathbf{y}_{DL}$ is bounded in $[0,1]$.
We empirically find these simple modifications on the loss function significantly improves the performance. 
We conjecture the unsuppressed biased output $\mathbf{y}_b$ allows our target model to better consider the \emph{intensity} of the bias, leading to a more robust target model.
In addition, when we train the target model, we empirically find a trend trend where instead of using the noise inputs as in $F_{b,G}(\mathbf{z},\mathbf{q})$, using the real images as such $F_{b}(\mathbf{v},\mathbf{q})$, is more effective, hence we use the output of $F_{b}(\mathbf{v},\mathbf{q})$ to train our target model. When the bias model is trained, it is trained with a noise vector to hallucinate the possible biases when only given the question, however, when we give it the real image, we find from~\Fref{fig:qual} that the output changes more drastically.

\section{Experiments}
\label{sec.experiments}

\begin{table*}[t]
\centering
\resizebox{.97\linewidth}{!}{
\begin{tabular}{l c c cccc c cccc}
% \midrule
\toprule
\multirow{2}{*}{Method} & &\multirow{2}{*}{Base} & \multicolumn{4}{c}{VQA-CP2 test} & & \multicolumn{4}{c}{VQA-CP1 test}\\  
\cmidrule{4-7} \cmidrule{9-12}
~ & ~ & ~ & {All} & {Yes/No} &  {Num} & {Other} & ~ & {All} & {Yes/No} & {Num} & {Other} \\
\bottomrule
SAN~\cite{yang2016stacked}  &  &- &   24.96 & 38.35 & 11.14  & 21.74 & & 32.50 & 36.86 & 12.47 & 36.22 \\
GVQA~\cite{agrawal2018don}   &  & -&  31.30 & 57.99 & 13.68  & 22.14 & & 39.23 & 64.72 & 11.87  & 24.86 \\
S-MRL~\cite{cadene2019rubi} & & -&  38.46 & 42.85 & 12.81 & 43.20 & & 36.38 & 42.72 & 12.59 & 40.35 \\
UpDn~\cite{anderson2018bottom}  &  &- & 39.94&42.46&11.93&45.09 & & 36.38 & 42.72 & 42.14 & 40.35
\\
\midrule
\multicolumn{12}{l}{\it Methods based on modifying language modules} \\
\midrule
DLR~\cite{jing2020dlr} & & UpDn & 48.87 & 70.99 & 18.72 & 45.57 & & -- & -- & -- & -- \\
VGQE~\cite{kv2020reducing} & & UpDn & 48.75 & -- & -- & -- & & -- & -- & -- & -- \\
VGQE~\cite{kv2020reducing} & & S-MRL & 50.11 & 66.35 & 27.08 & 46.77 & & -- & -- & -- & -- \\
\midrule
\multicolumn{12}{l}{\it Methods based on strengthening visual attention} \\
\midrule
HINT~\cite{selvaraju2019taking} &  & UpDn & 46.73 & 67.27 & 10.61 & 45.88 & & -- & -- & -- & -- \\
SCR~\cite{wu2019self} & & UpDn & 49.45 & 72.36 & 10.93 & 48.02 & & -- & -- & -- & -- \\
\midrule
\multicolumn{12}{l}{\it Methods based on ensemble models} \\
\midrule
AReg~\cite{ramakrishnan2018overcoming} & & UpDn &  41.17 & 65.49 & 15.48 & 35.48 & & 43.43 & 74.16 & 12.44 & 25.32 \\
RUBi~\cite{cadene2019rubi} & & UpDn& 44.23 & 67.05 & 17.48 & 39.61 & & 50.90 & 80.83  &  13.84  & 36.02 \\
LMH~\cite{clark2019dont} & & UpDn& 52.45 & 69.81 & \textbf{\underline{44.46}} & 45.54 & & 55.27 & 76.47 & \textbf{26.66} & \textbf{45.68} \\
CF-VQA(SUM)~\cite{niu2021cfvqa} & & UpDn & 53.55 & \textbf{\underline{91.15}} & 13.03 & 44.97 & & 57.03 & \underline{\textbf{89.02}} & 17.08 & 41.27 \\
CF-VQA(SUM)~\cite{niu2021cfvqa} & &  S-MRL & {55.05} & {90.61} & {21.50} & 45.61 & & \textbf{57.39} & \textbf{88.46} & 14.80 & 43.61 \\
CF-VQA(SUM)~\cite{niu2021cfvqa} + IntroD~\cite{niu2021introd} & &  S-MRL & {55.17} & \textbf{90.79} & {17.92} & 46.73 & & -- & -- & -- & -- \\
GGE~\cite{han2021greedy} & & UpDn& \textbf{57.32} & 87.04 & 27.75 & \underline{\textbf{49.59}} & & -- & -- & -- & -- \\

\midrule
\textbf{GenB (Ours)} & & UpDn & \underline{\textbf{59.15}}& 88.03 & \textbf{40.05} & \textbf{49.25} &  & \underline{\textbf{62.74}} & 86.18 & \underline{\textbf{43.85}} & \underline{\textbf{47.03}} \\
\bottomrule
\toprule
\multicolumn{12}{l}{\it Methods based on balancing training data} \\
\midrule
CVL~\cite{Abbasnejad_2020_CVPR} & & UpDn & 42.12&45.72&12.45&48.34 & --&--&--&--\\
RandImg~\cite{teney2020value} &  & UpDn &  55.37 & 83.89 & {41.60} & 44.20 & & -- & -- & -- & -- \\
SSL~\cite{zhu2021overcoming} &  & UpDn & {57.59} & 86.53 & 29.87 & 50.03
 & & -- & -- & -- & -- \\
CSS~\cite{chen2020counterfactual} & & UpDn & 58.95 & 84.37 & 49.42 & 48.21
 & & 60.95 & 85.60 & 40.57 & 47.03 \\
CSS~\cite{chen2020counterfactual} + IntroD~\cite{niu2021introd} & & UpDn & 60.17 & 89.17 & 46.91 & 48.62
 & & -- & -- & -- & -- \\

MUTANT~\cite{gokhale2020mutant} & & UpDn &  61.72 & 88.90 & 49.68 & 50.78 & & -- & -- & -- & -- \\
D-VQA~\cite{wen2021debiased}& & UpDn & 61.91 & 88.93&52.32&50.39 & &--&--&--&--\\
KDDAug~\cite{chen2022kddaug}& & UpDn & 60.24 & 86.13&55.08 &48.08 & &--&--&--&--\\
\bottomrule
\end{tabular}
}
\caption{Experimental results on VQA-CP2 and VQA-CP1 test set. \textbf{\underline{Best}} and \textbf{second best} results are styled in this manner within the column. We do not directly compare to methods that change the distribution of the training data as it does not go with the philosophy of VQA-CP~\cite{niu2021cfvqa} (which is to test whether VQA models rely on prior training data~\cite{agrawal2018don}). 
Among the compared baselines, our method GenB shows the best performance by a noticeable margin.}
% \vspace{-2mm}
\label{tab:overall}
\end{table*}

In this section, we explain our experimental setting and show our quantitative and qualitative experimental results.

\subsection{Setup}
\noindent\textbf{Dataset and evaluation metric.} We conduct our experiments within the VQA datasets that are commonly used for diagnosing bias in VQA models. In particular, we test on the VQA-CP2 and VQA-CP1 datasets~\cite{agrawal2018don} and the GQA-OOD dataset~\cite{kervadec2021gqaood}. 
For evaluation on all datasets, we take the standard VQA evaluation metric~\cite{antol2015vqa}. In addition to this, we also evaluate our method on the VQA-CE~\cite{dancette2021beyond}, which is based on the VQA v2 dataset, and is a newly introduced VQA evaluation protocol for diagnosing how reliant VQA models are on shortcuts.

\noindent\textbf{Baseline architecture.} 
% Following recent works, 
Unless otherwise stated, we adopt a popular VQA baseline architecture UpDn~\cite{anderson2018bottom} as both our ensemble bias model $F_b$ and our target model $F$. We list the details of the generator and discriminator in the supplementary material.
During training, we train both the bias model and target model together, then we use the target model only for inference. 

\subsection{Results on VQA-CP2 and VQA-CP1}

We compare how our method GenB performs in relation to the recent state-of-the-art methods that focus on bias reduction as shown in~\Tref{tab:overall}.
For VQA-CP2, we first list the \emph{baseline architectures} in the first section. Then, we compare our method to the 
methods that \emph{modify language modules} (DLR~\cite{jing2020dlr}, VGQE~\cite{kv2020reducing}),
\emph{strengthen visual attention} (HINT~\cite{selvaraju2019taking}, SCR~\cite{wu2019self}),
\emph{ensemble based methods}, (AReg~\cite{ramakrishnan2018overcoming}, RUBi~\cite{cadene2019rubi}, LMH~\cite{clark2019dont}, CF-VQA~\cite{niu2021cfvqa}, GGE~\cite{han2021greedy})
and \emph{balance training data} by changeing the training distribution (CVL~\cite{Abbasnejad_2020_CVPR}, RandImg~\cite{teney2020value}, SSL~\cite{zhu2021overcoming}, CSS~\cite{chen2020counterfactual}, Mutant~\cite{gokhale2020mutant}, D-VQA~\cite{wen2021debiased}, KDDAug~\cite{chen2022kddaug}) in the respective sections.
Among the balancing training data methods, while some methods swap the image and questions from the supposed pairs~\cite{teney2020value,zhu2021overcoming,wen2021debiased}, counterfactuals based methods generate counterfactual samples masking critical words or objects~\cite{chen2020counterfactual} or by using an external inpainting network to create a new subset of data~\cite{gokhale2020mutant}. In addition, as IntroD~\cite{niu2021introd} is seen as a technique built on top of certain previous methods, we include the scores in their respective categories. 
Our model (GenB) is in line with the ensemble models listed. 
Following previous works~\cite{niu2021cfvqa}, we do not compare to methods that change training distribution as these methods conflict with the original intent of VQA-CP~(which is to test whether VQA models rely on prior training data~\cite{agrawal2018don})
and are listed in the bottom of~\Tref{tab:overall}.

In \Tref{tab:overall}, our method achieves state-of-the-art performance on VQA-CP2, surpassing the second best (GGE~\cite{han2021greedy}) by 1.83\%. 
The performance of our model on all three categories (``Yes/No,'' ``Num,'' ``Other'')
are within top-3 consistently for the same backbone architecture.
Our method also performs highly favorably on ``Other''. 

We also show how our method performs on the VQA-CP1 dataset, which is a subset of the VQA-CP2 dataset.
Note that not all of the baselines are listed as we only list the scores that are available in the respective papers.
Our method also shows the state-of-the-art results on this dataset with a significant performance improvement over the second best among the methods compared, CF-VQA(SUM)~\cite{niu2021cfvqa}.
Even when compared to the available method CSS~\cite{chen2011collecting}, that we do not compare as it is considered unfair according to~\cite{niu2021cfvqa}, our method shows a significant gap in performance.
Compared to the second best method, CF-VQA(Sum)~\cite{niu2021cfvqa} on UpDn, our method improves the overall performance by 5.60\% while also having the best performance in both ``Num'' and ``Other'' category, by 3.28\% and 2.41\% performance improvements respectively.

\begin{table}[t]
    \centering
    \resizebox{.7\linewidth}{!}{
    \begin{tabular}{l c c c c}
        \toprule
        \multirow{2}{*}{Method} & \multicolumn{4}{c}{GQA-OOD Test}\\  
        \cmidrule{2-5}
         & All& Tail & Head & Avg \\
        \midrule
        UpDn~\cite{anderson2018bottom}&46.87&42.13&49.16&45.65 \\
        RUBi~\cite{cadene2019rubi}& 45.85 & 43.37 & 47.37 &45.37 \\
        LMH~\cite{clark2019dont}& 43.96 & 40.73 & 45.93 &43.33 \\
        CSS~\cite{chen2020counterfactual}&44.24&41.20&46.11&43.66 \\
        \midrule
        \textbf{GenB (Ours)}&\textbf{49.43}&\textbf{45.63}&\textbf{51.76} & \textbf{48.70} \\
        \bottomrule
         
    \end{tabular}
    }
    \caption{Experimental results on the GQA-OOD dataset. Our method shows the best performance in all the metrics compared to the state-of-the-art methods by a significant margin. The results show that our method robust on GQA-OOD as well.}
    \label{table:GQA-OOD}
\end{table}

\begin{table}[t]
    \centering
    \resizebox{.7\linewidth}{!}{
    \begin{tabular}{l c c c }
        \toprule
        \multirow{2}{*}{Method} & \multicolumn{3}{c}{VQA-CE}\\  
        \cmidrule{2-4}
        & Overall& Counter & Easy \\
        \midrule
        UpDn~\cite{anderson2018bottom}&{63.52}&33.91&{76.69} \\
        RUBi~\cite{cadene2019rubi}&61.88&32.25&75.03 \\
        LMH~\cite{clark2019dont}&61.15&34.26&73.13 \\
        RandImg~\cite{teney2020value}&63.34&34.41&76.21 \\
        CSS~\cite{chen2020counterfactual}&53.55&34.36&62.08 \\
        \midrule 
        \textbf{GenB (Ours)}&57.87&{34.80}&68.15 \\
        \bottomrule
         
    \end{tabular}
    }
    % \vspace{-2mm}
    \caption{Evaluation on the VQA-CE protocol. Ours shows the best performance in counterexamples (listed as Counter) which is the main scope of the VQA-CE.}
    \vspace{-2mm}
    \label{table:vqace}
\end{table}

\subsection{Results on GQA-OOD}

Recently, a new dataset for VQA debiasing, the GQA-OOD~\cite{kervadec2021gqaood} dataset, has been released and to test the debiasing ability of our method, we show our results in~\Tref{table:GQA-OOD}. 
We compare our method to available recent state-of-the-art ensemble based methods RUBi~\cite{cadene2019rubi}, LMH~\cite{clark2019dont}, and CSS~\cite{chen2020counterfactual}. 
Our method shows the best performance in all metrics compared to the state-of-the-art methods by a significant margin. Even when compared to methods that show similar performance to GenB in VQA-CP2 like CSS, GenB significantly outperforms it in GQA-OOD by 5.19\% in Overall.
Interestingly, although all of the listed previous methods outperform the base UpDn model in other datasets, they show a performance degradation on GQA-OOD. Unlike these methods, our method GenB is able to increase performance on GQA-OOD, showing the robustness of GenB.

\subsection{Results on VQA-CE}
We further evaluate our method on the newly introduced evaluation protocol that measures how much a VQA model depends on shortcuts called VQA-CounterExamples (VQA-CE)~\cite{dancette2021beyond} in~\Tref{table:vqace}.
This dataset is based on the VQA v2 dataset and lists three categories: Overall, Easy, and Counter; which are the total score on VQA v2 validation set, the subset of samples where the shortcuts of image/question give the correct answer, and the subset of samples where the shortcuts of image/question give incorrect answers respectively. 
Although the overall score for VQA-CE is dependent on the VQA v2 performance,
we show that our model shows the best performance on Counter, which is the main point of interest in this dataset.

\subsection{Ablation Studies}
Our method (GenB) includes several different components as shown from \Sref{sec:GenB} to \Sref{sec:debiasing}. To understand the effects of each component, we run an ablation study on the VQA-CP2 dataset. For all experiments, the results are of the target model and as the purpose of the bias model is to capture bias instead of correctly predicting the answers, we do not consider the predictions of the bias model. For all our ablation tables, we also add the UpDn baseline in the first row, the model in which our target model and bias model is based on for comparison. To further understand whether GenB can be applied to other networks, we further include an ablation study of GenB on other VQA architectures.

\begin{table}[t]
    \centering
    \resizebox{.9\linewidth}{!}{
    \begin{tabular}{l c c c c c }
        \toprule
        \multirow{2}{*}{Training Loss} &\multirow{2}{*}{Bias Model} & \multicolumn{4}{c}{VQA-CP2 test}\\
        \cmidrule{3-6}
         && All& Yes/No & Num & Other \\
        \midrule
        BCE&UpDn&39.94&42.46&11.93&45.09 \\
        \midrule
        BCE & GenB&56.98& 88.82 & 19.39 & \textbf{49.86}\\
        BCE + DSC& GenB&56.54& \textbf{89.06} & 21.29 & 49.79 \\
        BCE + Distill& GenB& 57.06 & 88.91 & 23.24 & 49.65\\
        BCE + DSC + Distill & GenB& \textbf{59.15} & 88.03 & \textbf{40.05} & 49.25 \\
        \bottomrule
         
    \end{tabular}
    }
    % \vspace{-2mm}
    \caption{We ablate the different losses we use to train the GenB bias model. All inferences scores are based on the target model except the first row. BCE loss is~\Eref{eq:bceloss}, which is the ground truth VQA loss. DSC refers to the discriminator loss~\Eref{eqn:genb} and Distill refers to the KL Divergence loss~\Eref{eqn:kl}. Although the DSC and Distill losses independently do not show large improvement, our final model with all losses show a large margin of improvement. 
    }
    \label{table:enstrain}
\end{table}

\noindent\textbf{Bias model training analysis.}
We ablate on the different losses we use to train the bias model from the target model and list our findings in~\Tref{table:enstrain} with BCE denoting the ground truth VQA loss~\Eref{eq:bceloss}, DSC denoting the discriminator loss~\Eref{eqn:genb}, and Distill denoting the KL Divergence distillation loss~\Eref{eqn:kl}. 
GenB trained on the BCE loss is already significantly helpful in debiasing the target model. 
Adding the DSC and Distill losses individually show similar performance to the GenB trained with BCE loss. However, by doubling down on the two losses, DSC and Distill, the model is better able to capture the biases held within the model, hence giving the significant performance boost. 
In addition, as the bias model and target model are trained concurrently, without a ground truth anchor (BCE loss), the model performs extremely poorly, and we conjecture that the bias model struggles to learn any meaningful bias. 
Note that BCE + DSC shows the best Yes/No score while our final model shows the best score in Num score. 
We conjecture that the bias model trained with our adversarial training framework is good at modeling the bias for simple answers like Yes/No while the bias with more complex answers like counting is hard to learn only with the adversarial training. 
We conclude that combining our knowledge distillation loss to the adversarial training is essential to learn the bias with complex answers including the Num category.

\begin{table}[t]
    \centering
    \resizebox{.9\linewidth}{!}{
    \begin{tabular}{l c c c c c}
        \toprule
        \multirow{2}{*}{Ensemble Debias Loss} &\multirow{2}{*}{Bias Model} & \multicolumn{4}{c}{VQA-CP2 test}\\  
        \cmidrule{3-6}
         & & All& Yes/No & Num & Other \\
        \midrule
        --& UpDn & 39.94&42.46&11.93&45.09 \\
        \midrule
        GGE~\cite{han2021greedy}  & UpDn& 47.40 & 64.45 & 13.96 & 47.64 \\
        Our Loss & UpDn & 52.47 & 88.20 & 30.09 & 40.38 \\
        \midrule
        RUBi~\cite{cadene2019rubi} & GenB & 30.77&72.78&12.15&13.87\\
        LMH~\cite{clark2019dont} & GenB & 53.99&75.89&\textbf{44.62}&45.08\\
        GGE~\cite{han2021greedy} &GenB& 49.51 & 70.63 & 14.08 & 48.16\\
        \midrule
        Ours Loss & GenB & \textbf{59.15} & \textbf{88.03} & 40.05 & \textbf{49.25}\\
        \bottomrule
         
    \end{tabular}
    }
    \caption{Ablation of ensemble debiasing loss functions and our debiasing loss function~\Eref{eqn:gradient}. Note that our loss improves the score of using the vanilla UpDn model when compared to GGE by a large margin. Note that GenB works best with our loss and shows a large performance improvement from GGE + GenB.}
    \label{table:lossabla}
\end{table}

\begin{table}[t]
    \centering
    \resizebox{.9\linewidth}{!}{
    \begin{tabular}{l c c c c}
        \toprule
        \multirow{2}{*}{Bias Model} & \multicolumn{4}{c}{VQA-CP2 test}\\  
        \cmidrule{2-5}
         & All& Yes/No & Num & Other \\
        \midrule
        UpDn&39.94&42.46&11.93&45.09 \\
        \midrule
        UpDn&52.47& 88.20 & 30.09 & 40.38\\
        Visual-Answer&41.03& 42.69 & 12.66 & 47.93\\
        Question-Answer&56.68& \textbf{89.30} & 20.78 & \textbf{49.43}\\
        GenB Visual&49.54& 72.05 & 12.58 & 47.89\\
        \textbf{GenB Question (Ours)}& \textbf{59.15} & 88.03 & \textbf{40.05} & 49.25 \\
        \bottomrule
         
    \end{tabular}
    }
    % \vspace{-2mm}
    \caption{Ablation of different bias models with our modified loss. We fix the target model architecture to be the UpDn model and show the performance of the target model by changing the bias model architecture between UpDn, Question-Answer model, Visual-Answer model, our GenB model but with image and generated noise, and our GenB model, which has question and generated noise. We show that our loss works with other networks, and although other networks can be helpful in debiasing, our model, GenB with question and noise is the best performing model by far.
    }
    \vspace{-2mm}
    \label{table:ensmods}
\end{table}

\noindent\textbf{Debiasing loss function analysis.} 
We list in~\Sref{sec:debiasing} the different losses we can use for debiasing and we show our results in~\Tref{table:lossabla}. We also list our new debiasing loss function~\Eref{eqn:gradient} and how it performs when a vanilla UpDn model is used instead of the GenB.

We analyze how GGE's loss function fairs when the vanilla UpDn model is used as a bias model. In the second and third row, although the GGE loss improves the score from the baseline by introducing a bias model, our loss exceeds the performance by 5.07\%. In the last two sections, we show the debiasing losses with the full GenB model with all training losses and other debiasing losses. Compared to GGE with GenB, our loss with GenB shows a 9.64\% performance gap. Given the GGE loss, the performance increase from the bias model is only 2.11\%, whereas our debiasing loss function is able to boost the score from the vanilla UpDn to GenB by 6.68\%. We conjecture that as our loss function accounts for the amount of bias a model captures, and the generative bias model is able to stochastically model the bias, our loss function captures the bias in a much more dynamic and helpful manner.

\begin{table}[t]
    \centering
    \resizebox{1\linewidth}{!}{
    \begin{tabular}{l c c c c c c}
        \toprule
        \multirow{2}{*}{Architecture} &\multirow{2}{*}{} & \multicolumn{4}{c}{VQA-CP2 test} & \multirow{2}{*}{$\Delta$ Gap}\\  
        \cmidrule{3-6}
         & & All& Yes/No & Num & Other \\
        \midrule
        UpDn~\cite{anderson2018bottom} & & 39.94 & 42.46 & 11.93 & 45.09 & \multirow{2}{*}{\textbf{+19.21}}\\
        UpDn~\cite{anderson2018bottom} + GenB & & \textbf{59.15} & \textbf{88.03} & \textbf{40.05} & \textbf{49.25} & \\
        \midrule 
        BAN$ ^\dag $ ~\cite{kim2018bilinear} & & 37.35 & 41.96 & 12.08 & 41.71 & \multirow{2}{*}{\textbf{+20.02}}\\
        BAN$ ^\dag $ ~\cite{kim2018bilinear} + GenB & & \textbf{57.37} & \textbf{89.11} & \textbf{29.52} & \textbf{48.37} & \\
        \midrule 
        SAN$ ^\dag $ ~\cite{yang2016stacked} & & 38.65 & 40.59 & 12.98 & 44.67 & \multirow{2}{*}{\textbf{+18.07}}\\
        SAN$ ^\dag $ ~\cite{yang2016stacked} + GenB & & \textbf{56.72} & \textbf{88.84} & \textbf{19.04} & \textbf{50.22} & \\
        \midrule
        LXMERT~\cite{lxmert2019} & & 46.23 & 42.84 & 18.91 & 55.51 &  \multirow{2}{*}{\textbf{+24.93}}\\
        LXMERT~\cite{lxmert2019} + GenB \textbf{(Ours Best)} &  & \textbf{71.16} & \textbf{92.24} & \textbf{64.71} & \textbf{61.89} \\ 
        \midrule
        Reported LXMERT Performance &&&&&\\
        \midrule
        LXMERT~\cite{lxmert2019} + MUTANT~\cite{gokhale2020mutant}&  & 69.52 & 93.15 & 67.17 & 57.78 & \\
        LXMERT~\cite{lxmert2019} + D-VQA~\cite{wen2021debiased} & & 69.75 & 80.43 & 58.57 & 67.23 & \\
        LXMERT~\cite{lxmert2019} + SAR~\cite{si2021check} & &  62.12 & 85.14 & 41.63 & 55.68 & \\
        \bottomrule
         
    \end{tabular}
    }
    \caption{Ablation of GenB on different architectures. $ ^\dag $ signifies our re-implementation. We find that adding GenB to other backbones show consistent improvement, showing that GenB is applicable on other baselines. Note that we include the reported LXMERT based performance in the last 4 rows. LXMERT + GenB is state-of-the-art among all reported VQA-CP2~\cite{wen2021debiased,si2021check,gokhale2020mutant}.}
    \label{table:arch}
    \vspace{-2mm}
\end{table}

\begin{figure*}
    \centering
    \includegraphics[width=.97\linewidth]{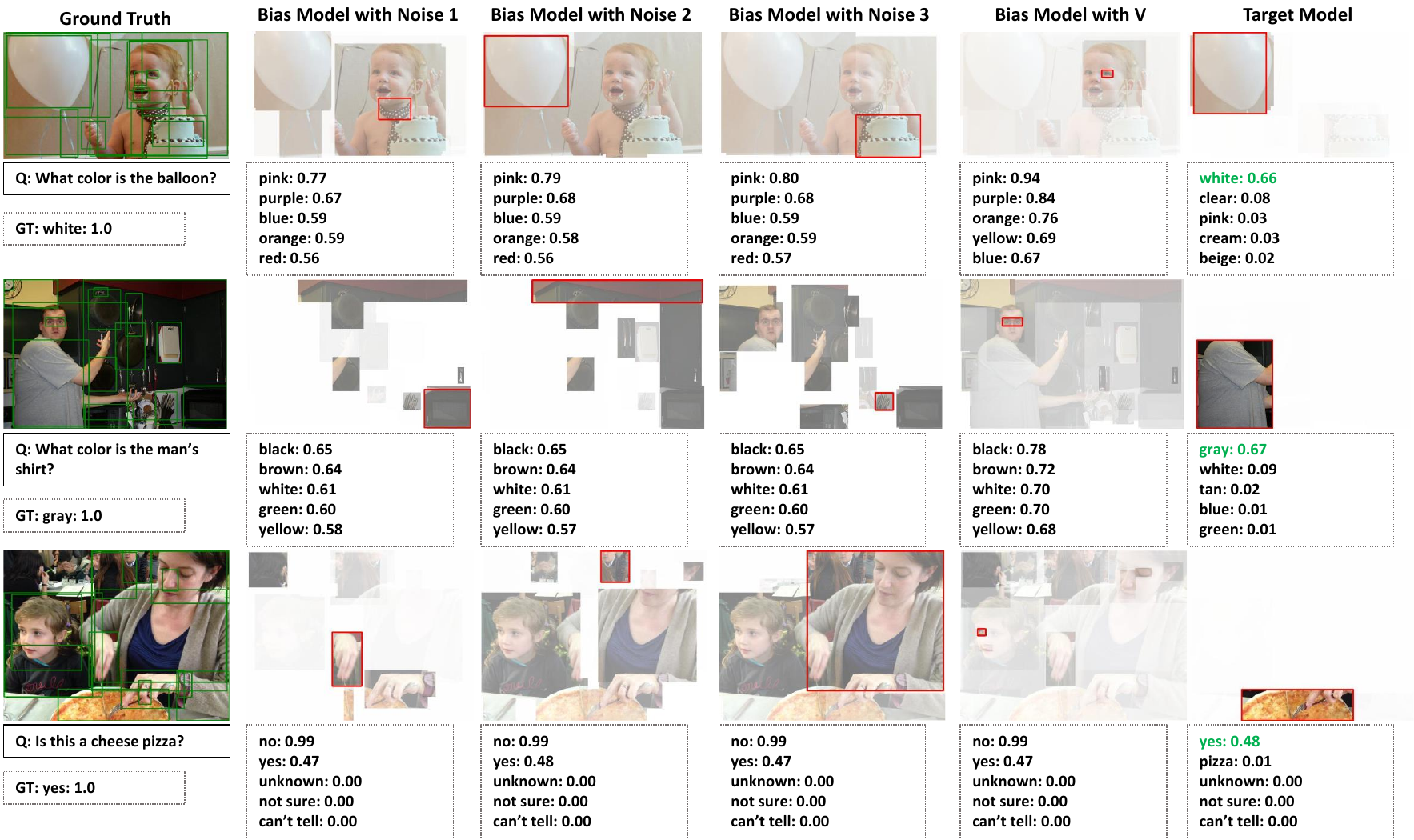}
    \caption{Qualitative results showing the predictions and attention scores of our target model and the bias model. We show the attention of the bias model with different noise to see where it would attend to. We also input the real image into the bias model to visualize the attention. The bias model's attention changes each time and does not attend to the critical areas of the image. 
    }
    \label{fig:qual}
\end{figure*}

\noindent\textbf{Effect of different bias models.}
We run an ablation experiment to understand the effects of different bias models and the addition of the Generator $G(\cdot)$ and its effects on the performance of the target model in~\Tref{table:ensmods}. 
We fix the target model as the UpDn model and show the performance of the target model by changing the bias model.
Note that the results are from the predictions of the target model, not the bias model, and the top row is the UpDn baseline.
We include the UpDn model as a bias model, the Visual-Answer model, which is a model that takes an image in as its input and predicts the answer, the Question-Answer model, which takes in a question and predicts the answer, GenB Visual model, which takes in an image together with noise to predict the answer, and our model GenB Question, which is the model we have shown up until this point. 
These models are used to debias the target model with our modified debiasing loss function.
We show that even adding noise to the image can be helpful, however, we find that our final model, the model with question and noise, outperforms other methods significantly.

\noindent\textbf{Effect on different backbones.} 
We further experiment on the robustness of our method GenB and its applicability to other VQA backbones (SAN~\cite{yang2016stacked}, BAN~\cite{kim2018bilinear}, and LXMERT~\cite{lxmert2019}) as shown in \Tref{table:arch}. For each architecture, we set both the target model and GenB model to be the same architecture. We show that GenB consistently shows significant performance improvements on all backbones, signifying that GenB is not simply limited to the UpDn setting. Note that LXMERT + GenB shows the highest LXMERT performance as reported in~\cite{wen2021debiased,si2021check,gokhale2020mutant}, which is the absolute state-of-the-art VQA-CP2 score so far. 
We believe this opens up a possibility for future works where different backbones can be mixed and matched for further debiasing.

\subsection{Qualitative Results}

We visualize the attention and predictions of our target and bias models in~\Fref{fig:qual}. We run the bias model three times with different noise to show the varying attention and biased predictions. 
% Note that since we used the noise input, the random attention is expected.
We also input the corresponding image with the question to see where the model would attend, and find that the attention is once again random as the image is previously unseen. As expected, the bias model's predictions change each time. Even with the same question type (\eg, what color), we find the biased model's predictions are noticeably different.
% Nevertheless, as expected, we see that the biased model shows a random attention mask. 
We find the target model is able to correctly attend to the salient region while predicting the correct answer by virtue of the bias model.
% , but the biased model outputs random attention scores with wrong predictions.

\section{Conclusion}
\label{sec.conclusion}

In this paper, we started with the intuition ``the better the bias model, the better we can debias the target model. Then how can we best model the bias?''
In response, we present a simple, effective, and novel generative bias model that we call GenB.
We use this generative model to learn the bias that may be inhibited by both the distribution and target model with the aid of generative networks, adversarial training, and knowledge distillation. In addition, in conjunction with our modified loss function, our novel bias model is able to debias our target model, and our target model achieves state-of-the-art performance on various bias diagnosing datasets and architectures.

\section*{Acknowledgements}
This work was supported by Institute of Information \& communications Technology Planning \& Evaluation (IITP) grant funded by the Korea government(MSIT) (No.2020-0-01373, Artificial Intelligence Graduate School Program(Hanyang University))

%%%%%%%%% REFERENCES
{\small
\bibliographystyle{ieee_fullname}
\bibliography{egbib}
}

\end{document}